\documentclass[a4paper, 11pt]{article}

\usepackage[english]{babel}
\usepackage[utf8x]{inputenc}
\usepackage[T1]{fontenc}
\usepackage{times}
\usepackage{latexsym}
\usepackage{microtype}
\usepackage{graphicx}
\usepackage[colorinlistoftodos]{todonotes}
\usepackage[margin=0.8in, top=1in]{geometry}
\usepackage{wrapfig,lipsum,booktabs}
\usepackage{subfig}
\usepackage{multirow}
\title{\vspace{-3em}Depthwise Separable Convolutions with Deep Residual Convolutions}
\author{
Md. Arid Hasan and Krishno Dey\\
\texttt{\{arid.hasan, krishno.dey\}@unb.ca} \\
\texttt{University of New Brunswick} \\
}
\date{}

\begin{document}
\maketitle

%\begin{abstract}
%\textit{\textbf{Abstract -- }}
%\end{abstract}
%\begin{abstract}
\vspace{-3em}
\section*{ABSTRACT}
\vspace{-1em}
The recent advancement of edge computing enables researchers to optimize various deep learning architectures to employ them in edge devices. In this study, we aim to optimize Xception architecture which is one of the most popular deep learning algorithms for computer vision applications. The Xception architecture is highly effective for object detection tasks. However, it comes with a significant computational cost. The computational complexity of Xception sometimes hinders its deployment on resource-constrained edge devices.
To address this, we propose an optimized Xception architecture tailored for edge devices, aiming for lightweight and efficient deployment. We incorporate the depthwise separable convolutions with deep residual convolutions of the Xception architecture to develop a small and efficient model for edge devices. The resultant architecture reduces parameters, memory usage, and computational load.
The proposed architecture is evaluated on the CIFAR 10 object detection dataset. The evaluation result of our experiment also shows the proposed architecture is smaller in parameter size and requires less training time while outperforming Xception architecture performance.
%\end{abstract}

\section*{KEYWORDS}
\vspace{-1em}
Depthwise Separable Convolution, Deep Residual, Xception

\section{Introduction}

Xception networks have demonstrated remarkable performance in computer vision and classification tasks, merging the ideas of Inception network and ResNet, but it replaces the inception modules with a depthwise separable convolution layer \cite{Xceptionnet}. While a regular convolutional layer uses filters that try to simultaneously capture spatial patterns (e.g., an oval) and cross-channel patterns (e.g., mouth + nose + eyes = face), a separable convolutional layer makes the strong assumption that spatial patterns and cross-channel patterns can be modeled separately. Thus, it is composed of two parts: the first part applies a single spatial filter for each input feature map, and the second part looks exclusively for cross-channel patterns. By employing separable convolutional layers, they reduce parameters, memory usage, and computational load compared to conventional convolutional layers, and in general, they even perform better.

Current algorithms in computer vision often face challenges related to computational resources and model complexity, particularly when deployed on edge devices with limited processing power and memory. Moreover, the increasing demand for lightweight and energy-efficient models exacerbates these issues, as existing architectures may not be optimized for such constraints. As a result, there's a pressing need to develop and adapt algorithms that strike a balance between accuracy and efficiency, ensuring seamless deployment on edge devices while maintaining robust performance in various computer vision tasks.

Our objective in this study is to further refine the Xception network for edge devices by optimizing its architecture to be more lightweight, ensuring it can run efficiently on such devices without significant compromise in model efficiency. %The Xception architecture consists of three main flows: Entry flow, middle flow, and exit flow, which we modify to adapt the network for edge devices.

The rest of the report structure is as follows: we discuss the deep learning models and limitations in Section \ref{sec:related}. We provide the details of our proposed architecture in Section \ref{fig:architecture}. Experimental details are discussed in Section \ref{sec:experiment}. We provide the detailed analysis and performance differences in Section \ref{sec:result}. Finally, we concluded our study in Section \ref{sec:conclusion}.
\section{Related Work}
\label{sec:related}

%brief introduction, why we need to use ML models on the edge
Artificial Intelligence(AI), Machine Learning(ML), and Deep Learning(DL) have recently advanced enormously. The world has witnessed a drastic rise in the usage of AI and ML across many domains, due to such advancements. Edge computing also saw a significant increase in the usage of AI and ML models. Training model centrally using aggregated data from edge devices ensures better prediction accuracy but involves several overheads such as data transfers, and model sharing with edge devices. %These overheads can be addressed by training models on edge devices, at the expense of a slight decrease in the prediction performance. Utilization of ML models at the edge addresses the need for low latency, efficient use of bandwidth, enhanced privacy and security, scalability, energy efficiency, and adaptability to the constraints of edge devices. However, edge devices usually have very minimal computational power,  which makes it challenging to train computationally extensive DL models on edge devices. 

% We are going to give a little background about the usage of ML models on edge
\subsection{AI Models for Edge Networks}
Several researchers have used ML models on edge devices. Studies such as \cite{murshed2021machine, merenda2020edge, verhelst2020machine} highlight current trends in edge computing and state-of-the-art ML and DL models used to train models on edge devices. Most of the studies use very lightweight ML models due to the resource limitation of edge devices. 
% Traditional machine learning
During the initial stages of edge computing development, researchers commonly utilized traditional machine learning models \cite{yazici2018edge, sudharsan2020edge2train}. As the field evolved, there was a discernible trend towards the incorporation of more sophisticated ML and DL models tailored for edge computing \cite{ding2019tdd, zhang2018improved}.  
%use of federated learning to use deep learning
Moreover, A survey conducted by Lim et al. demonstrates the use of the federated learning concept to train deep learning models on edge devices \cite{9060868}.  Over time, edge devices are getting more sophisticated, and we're seeing more efficient models emerging. This is paving the way for a new era in edge computing.

%conclusion of this section
AI is undergoing continuous advancement, leading to increased sophistication in ML and DL models. This progress is accompanied by the emergence of new algorithms, contributing to the complexity and resource requirements of these models. Training such advanced models on low-resourced edge devices poses a significant challenge. Optimizing these models to minimize complexity and resource consumption is imperative for enhancing their feasibility on edge devices. This optimization can significantly expand the practical usage of these models in edge computing scenarios.

%optimized models on edge 
\subsection{Optimized Machine Learning Models for Edge Networks}
Several studies have been completed to optimize machine learning models for edge devices. The early studies only include reducing the number of layers from the machine learning models. The issue with reducing layers also harms the performance.
%ali
The study of Runwei et al. \cite{ding2019tdd} introduced a Tiny defect detection (TDD) network based on R-CNN, aiming to detect small defects in PCBs, overcoming challenges like tiny defect sizes and limited data. They addressed these issues with three innovations: designing reasonable anchors using k-means clustering, enhancing feature map relationships across levels for tiny defect detection, and employing online hard example mining to improve ROI proposal quality and dataset utilization. Their method surpassed other approaches in performance.

Zhang et al. \cite{zhang2018improved} used transfer learning and data augmentation to train a deep CNN for defect detection, addressing data scarcity by leveraging the lower layers of VGG-16 and augmenting the dataset with artificial defect data and affine transformations. After fine-tuning, they employ a sliding window approach for defect localization, showing superior performance compared to traditional shallow feature-based methods in PCB defect detection.
In \cite{kumar2020energy} introduce a plugin designed to enhance the energy efficiency of Java-based machine learning code. This plugin offers suggestions to improve energy consumption across various aspects of Java programming, including data types, operators, control statements, Strings, exceptions, objects, and Arrays. Their assessment reveals a significant 14.46\% reduction in energy usage when applied to optimize the machine learning software WEKA, with a minimal 0.48\% decrease in accuracy.

%parisa
This study by \cite{kristiani2020isec} introduced L-CNN, a Lightweight Convolutional Neural Network for efficient human object detection in edge-based smart surveillance. Leveraging depthwise separable convolution and SSD, it addresses computational challenges on resource-limited edge devices. Focused on real-time pedestrian detection, L-CNN is trained on ImageNet and VOC07 datasets, implemented on a Raspberry Pi 3 using openCV. Anticipated outcomes include reduced computation workload, real-time performance, and improved handling of complex scenarios, contributing to the advancement of edge computing in smart surveillance. Nikouei et al. \cite{nikouei2018real} proposed an optimized deep learning model for edge-based image classification, utilizing techniques like image preprocessing and CPU optimization. InceptionV3 is chosen for edge deployment, and additional optimization using a Model Optimizer enhances accuracy and reduces loading time and model size. Experiments demonstrate improved performance after fine-tuning, with VGG16 identified as the most accurate and efficient model. Future work is recommended for testing in diverse cases, exploring GPU performance, addressing edge computing security, and evaluating alternative edge environments. Another study by Alzubi et al. \cite{alzubi2022optimized} introduced ESOML-IDS, a novel intrusion detection model for Fog Computing (FC) and Edge Computing (EC). Utilizing Effective Seeker Optimization (ESO) and Comprehensive Learning Particle Swarm Optimization (CLPSO) with Denoising Autoencoder (DAE), it addresses challenges like limited resources and large-dimensional features. Experimental results show ESOML-IDS outperforms recent approaches in accuracy, precision, recall, and F1 score, demonstrating its potential for enhancing cybersecurity in FC and EC environments.

%other two paper
Mih et al. \cite{nuh2023developing} present modified Xception a novel lightweight architecture for training Xception on resource constraint edge devices. The performance of the optimized Xception model was evaluated on the defect detection task, the performance suggests the optimized architecture exhibits better performance than existing lightweight models in terms of accuracy, memory usage, GPU utilization, and power consumption.

\subsection{Limitation of Current ML Models and Proposal}
In recent years edge computing gained attention from researchers and researchers have been trying to incorporate deep learning architecture to train on edge devices since then. Deep learning models such as XceptionNet\footnote{Has millions of trainable parameters} consume large computational resources to train and infer. Despite the limited computational resources such as memory, CPU, etc., researchers have been trying to optimize these models without compromising the performances \cite{nuh2023developing}.

In this study, we mainly focus on the challenges of optimizing the deep learning models for edge devices that have been discussed in the previous subsections. %Our primary goal is to reduce the size of the deep learning model networks and train the networks on edge devices that will reduce memory and CPU consumption during training without compromising performance. In this study, we will also reduce some less important layers that have less effect on performance and also reduce the filtering computation.

\section{Proposed Architecture}
\label{sec:architecture}

We propose a modified Xception architecture that relies on convolutional neural network architecture based entirely on depthwise separable convolution layers. On top of that, we employed deep residual on a few layers and added two fresh layers of convolutional neural networks based on entirely deep residual connections.

\begin{figure*}[!ht]
    \centering
  \subfloat[Entry Flow]{%
       \includegraphics[width=0.25\linewidth]{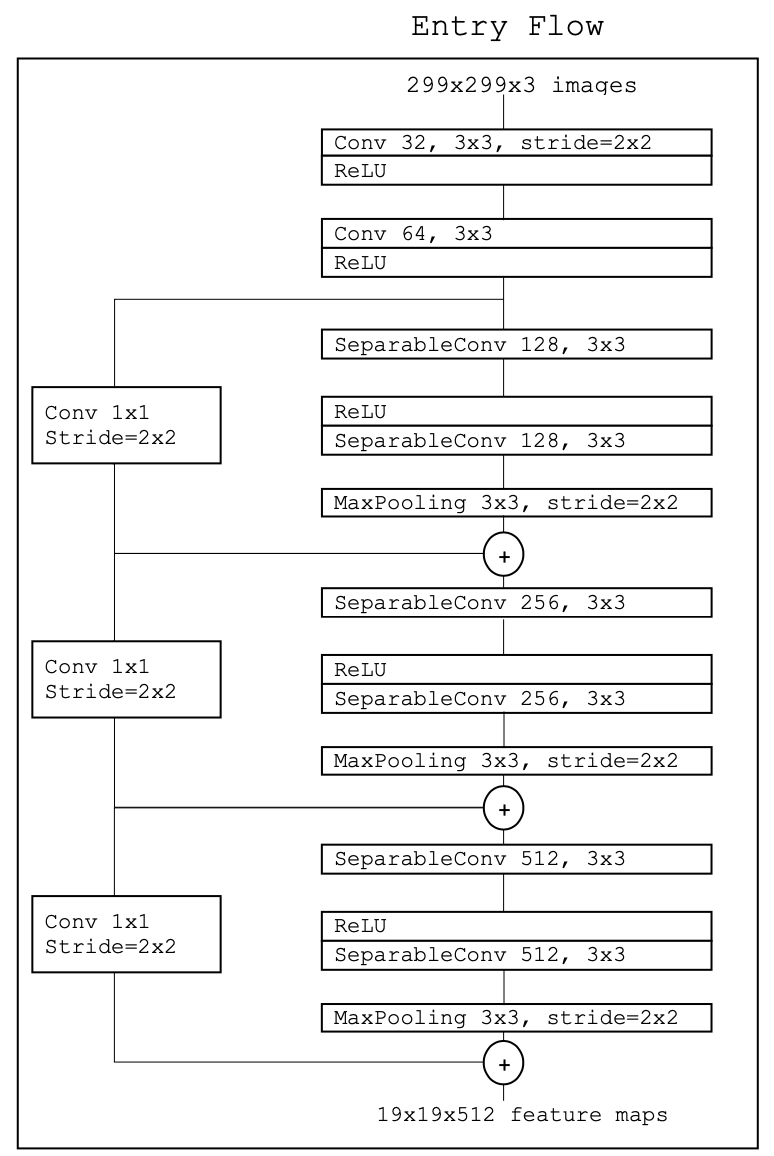}}
    %\hfill
  \subfloat[Middle Flow 1]{%
        \includegraphics[width=0.25\linewidth]{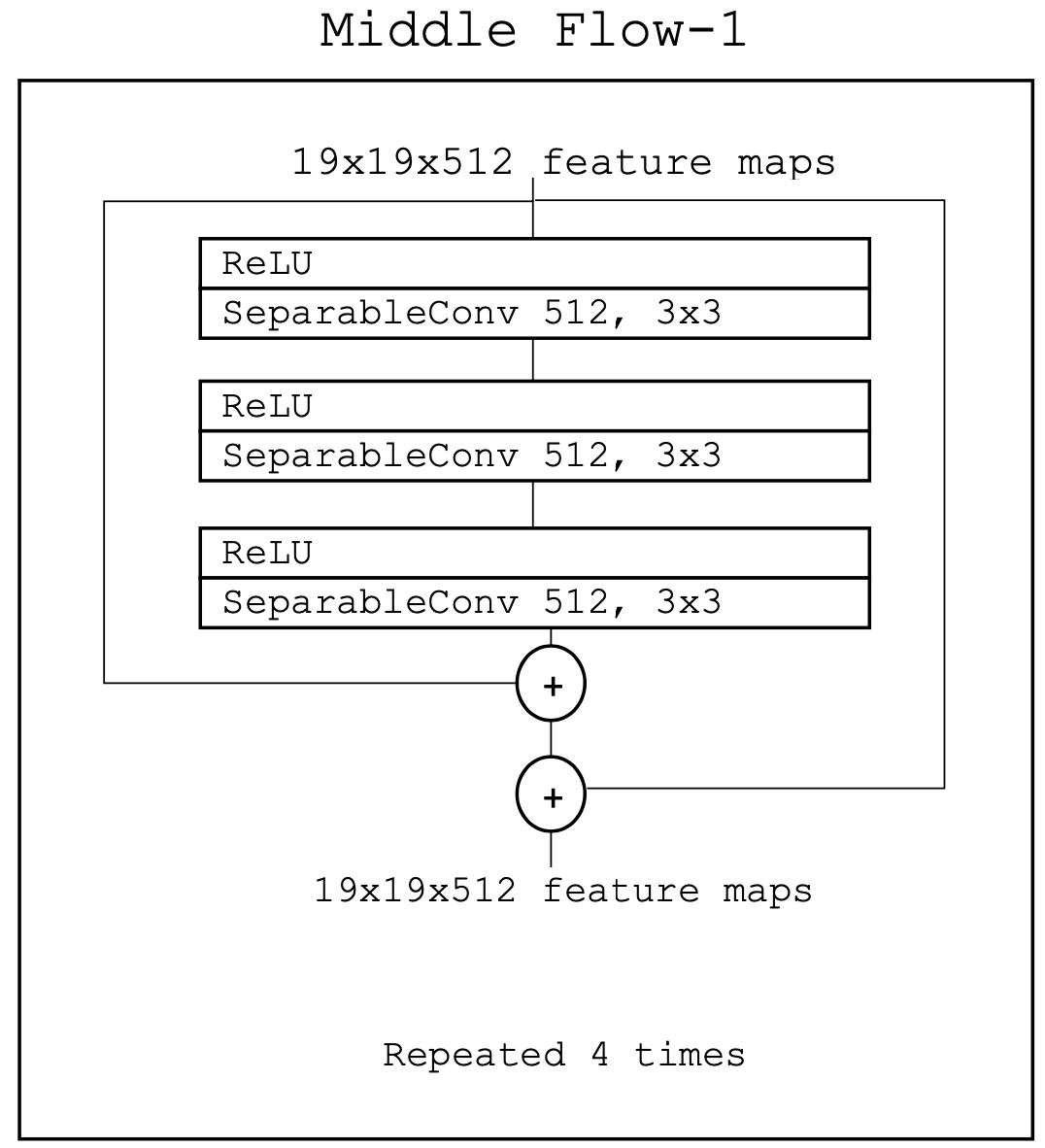}}
    %\\
  \subfloat[Middle Flow 2]{%
        \includegraphics[width=0.25\linewidth]{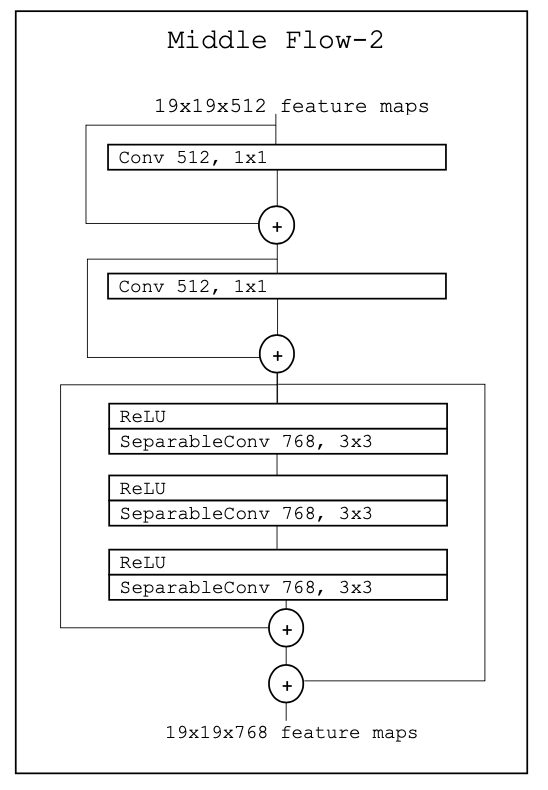}}
    %\hfill
  \subfloat[Exit Flow]{
        \includegraphics[width=0.25\linewidth]{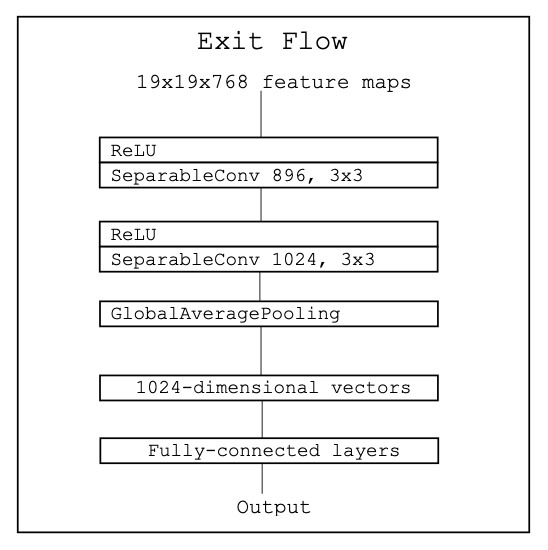}}
  \caption{Proposed Architecture.  the data first goes through the entry flow, then through the middle flow-1 which is repeated 4 times, then through the middle flow-2, and finally through the exit flow. Note that all Convolution and SeparableConvolution layers are followed by batch normalization (not included in the diagram). All SeparableConvolution layers use a depth multiplier of 1 (no depth expansion).}
  \label{fig:architecture}
\end{figure*}

We provide the detailed architecture of our proposed modified Xception architecture in Figure \ref{fig:architecture}. The proposed architecture has 26 convolutional layers, forming the feature extraction base of the network. In our experimental evaluation we will exclusively investigate image classification and therefore our convolutional base will be followed by a logistic regression layer. Moreover, we inserted fully connected layers before the logistic regression layer. The 26 convolutional layers are structured into 12 modules, all of which have linear residual connections around them while the middle flow-1 and the last module of middle flow-2 have two linear residual blocks, except for the first and last modules. In short, the proposed architecture is a linear stack of depthwise separable convolution layers with residual connections. This makes the architecture very easy to define and modify.
%\resizebox{.9\linewidth}{!}{
%}

% \begin{figure}[hbt!]
% \scalebox{\textwidth}{
%     \centering
%     \begin{subfigure}[t]{0.3\textwidth}
%         \centering
%         \includegraphics[width=3.5in]{figures/entry_flow.pdf}
%         %\caption{Entry Flow}
%     \end{subfigure}%
%     %~ 
%     \begin{subfigure}[t]{0.29\textwidth}
%         \centering
%         \includegraphics[width=3.5in]{figures/midflow-1.pdf}
%         %\caption{Mid Flow}
%     \end{subfigure}
%     %~
%     \begin{subfigure}[t]{0.4\textwidth}
%         \centering
%         \includegraphics[width=3.5in]{figures/midflow-2.pdf}
%         %\caption{Exit Flow Flow}
%     \end{subfigure}
%     \caption{Proposed Architecture.  the data first goes through the entry flow, then through the middle flow-1 which is repeated 4 times, then through the middle flow-2, and finally through the exit flow. Note that all Convolution and SeparableConvolution layers are followed by batch normalization (not included in the diagram). All SeparableConvolution layers use a depth multiplier of 1 (no depth expansion).}
%     \label{fig:architecture}
% }
% \end{figure}

\section{Experimental Details}
\label{sec:experiment}

In this section, we first discuss the CIFAR-10 dataset followed by experimental details of our study.

\subsection{Dataset}
\textbf{CIFAR-10:} The CIFAR 10 \cite{alex2009learning} dataset was used for the evaluation of these models. CIFAR 10 and CIFAR 100 are one of the most accepted object detection benchmarks to evaluate computer vision models. CIFAR 10 consists of 60000 32x32 color images in 10 classes, with 6000 images per class. The dataset is divided into 50000 training images and 10000 test images. CFIAR 100 is identical to the CIFAR-10, except it has 100 classes containing 600 images each. In this study, we utilized the CIFAR-10 dataset due to resource constraints.

\subsection{Hyperparameters}
We keep the same hyperparameters across the different experimental settings for fair comparison. We use \textit{CrossEntropyLoss} as the loss function for all the experiments. Optimizer \textit{Adam} is used with a learning rate of $2e-5$. Finally, we train 10 epochs in all experiments and generate results for every epoch to compare the performances across the model.

\subsection{Experiment Settings}

We conducted our experiments using three different settings. We provide the details of our experiment settings below.

\paragraph{XceptionNet Setting} In this experiment setting, we used the Xception architecture to train and evaluate the model using the CIFAR-10 dataset. We did not modify the layers or dimensions of layers in this experiment.

\paragraph{Optimized Setting} In this experiment setting, we used our proposed architecture of depthwise separable convolutions with deep residual convolutions. We train and evaluate the CIFAR-10 dataset on the proposed architecture using the previously defined hyperparameters.

\paragraph{Optimized with Data Setting} In this experiment setting, we first processed the data. To process the data, we average the previous, current, and next pixel values to create a better border for each object. After processing the data, we train and evaluate on proposed architecture using the same hyperparameter settings.

\section{Model Comparison and Discussion}
\label{sec:result}

In this section, we present the training and evaluation results of our proposed optimization of Xception. We compare the result of the proposed optimized models (i.e., optimized XceptionNet, and optimized XceptionNet with data) with the original Xception\cite{Xceptionnet} architecture. 
%dataset

\subsection{Trainable Parameters}

The Xception architecture has a total of 20.83 million trainable parameters while our proposed optimized architecture has only 7.43 million trainable parameters. Our proposed architecture has approximately 3 times fewer trainable parameters than the original Xception architecture which makes the model more lightweight. Moreover, due to less trainable parameters, our proposed architecture provides better performances in various criteria such as training time, memory consumption, and validation time compared to Xception architecture. The detailed comparison of performances among different criteria is discussed in the following sub-sections. 

\subsection{Training Time and Loss}
 
\begin{figure}[!htb]
\centering
\subfloat[Training time.]{\includegraphics[width=0.49\textwidth]{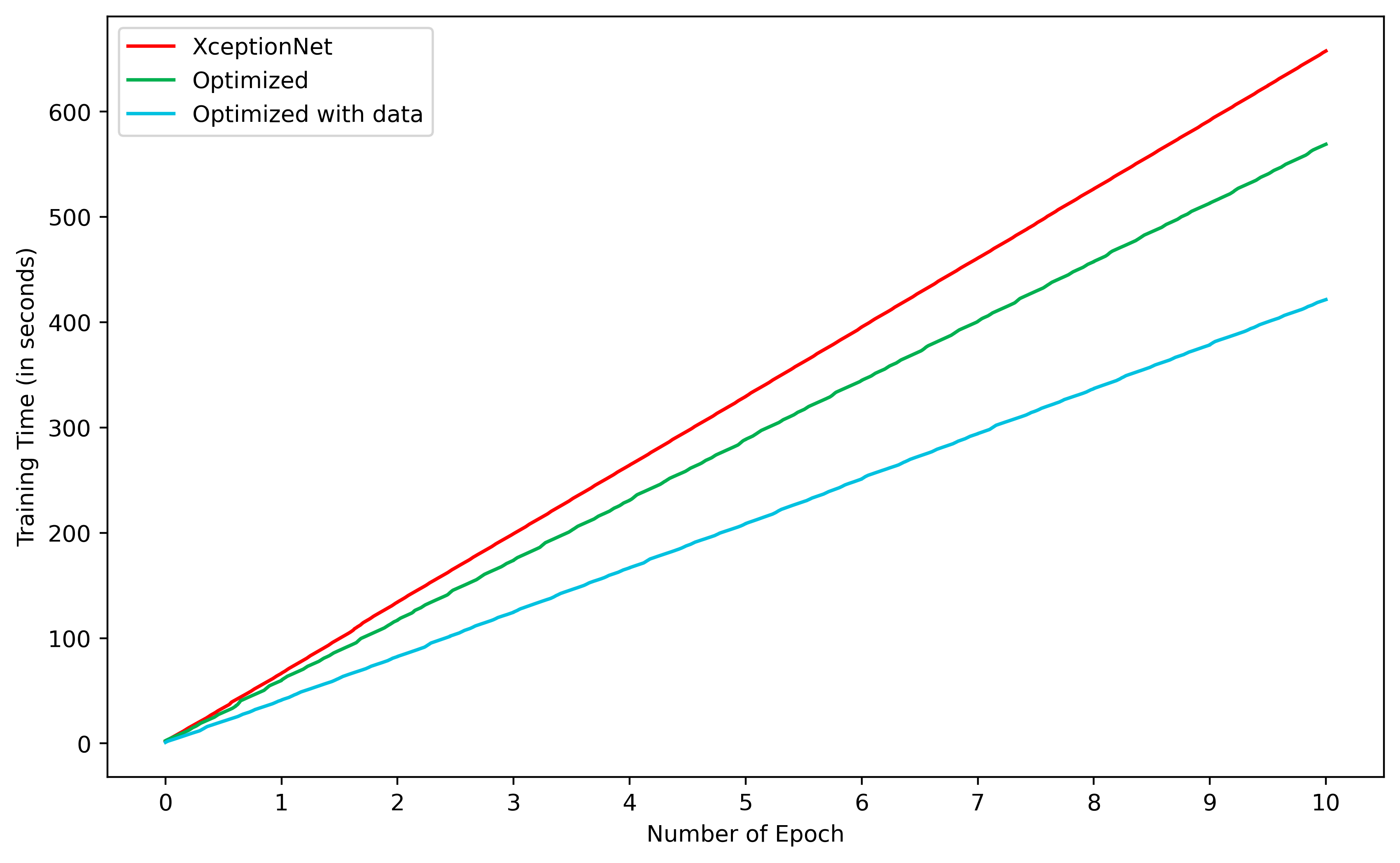}}%
\hfill 
\subfloat[Training loss.]{\includegraphics[width=0.49\textwidth]{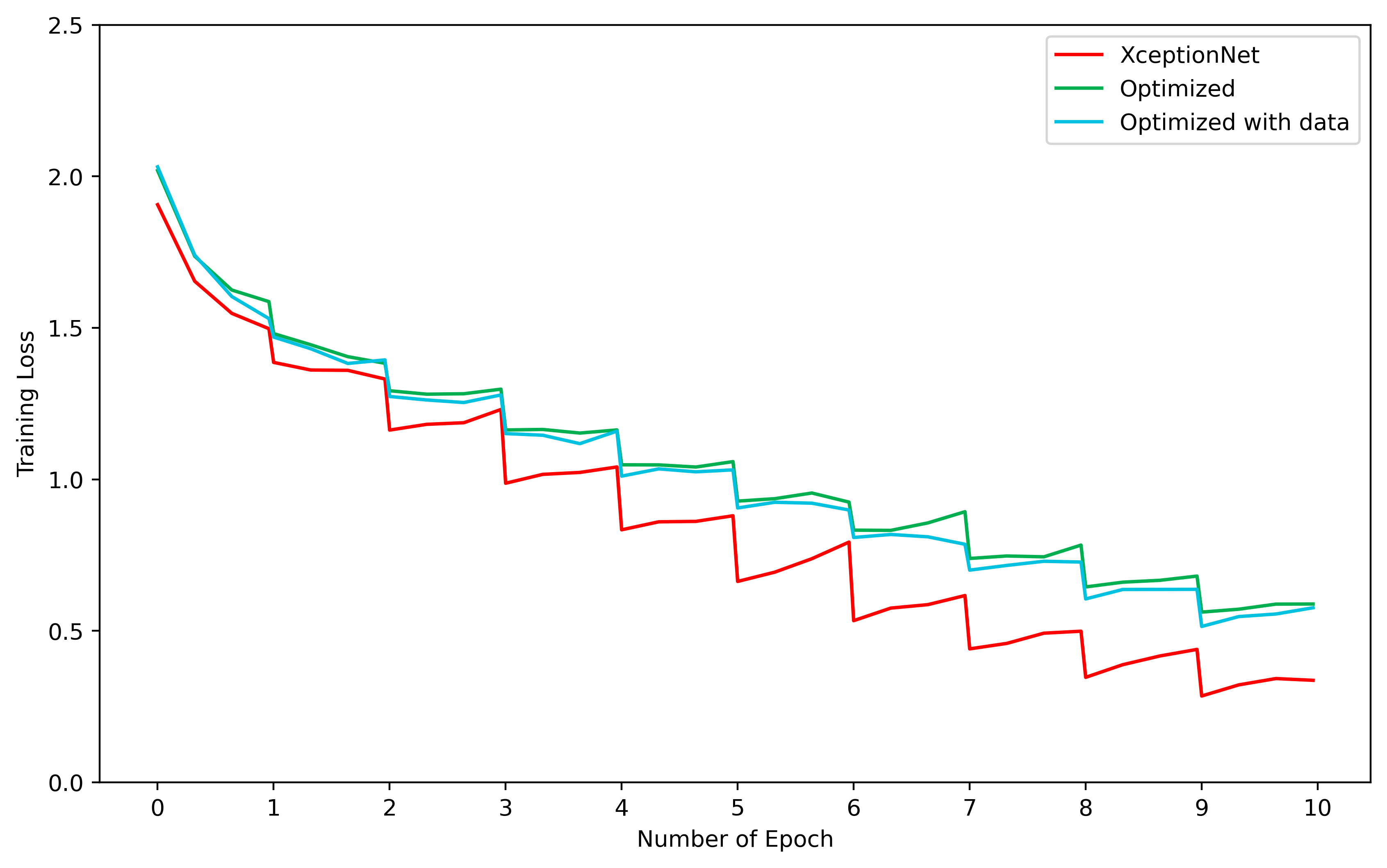}}%
\caption{\textbf{(a):} Comparison of training time among XceptionNet, optimized XceptionNet, and optimized XceptionNet with data. \textbf{(b):} Comparison of training loss among XceptionNet, optimized XceptionNet, and optimized XceptionNet with data.}
\label{trainging-time-loss}
\end{figure}

Since every model was generalizing over the course of the ten epochs, we ran every model for ten echos. First, We train the original XceptionNet with the CIFAR 10 dataset. Then we use the same data to evaluate the optimized XceptionNet. Then we evaluate the optimized XceptionNet with optimized input data, i.e., we used the average pixel values of images during the training process. While calculating the output of the convolution operation on the image, we average the pixel values of the three rows (current row, row above the current, and row below the current row). The average calculation during the convolution process further reduces the computation complexity of our optimized XceptionNet without compromising the performance to a great degree, which is evident in the evaluation result. Such down-scaling of computation complexity makes our optimized XceptionNet model suitable for edge devices.

%training time
Figure 2(a) shows the time taken to train the models on the CIFAR 10 dataset for 10 epochs. The graph suggests that optimized XceptionNet requires less training time than the original XceptionNet architecture. Moreover, reducing the dimension of images by averaging the pixel values (optimized Xception with data) reduces the model size and takes less time to train compared to optimized XceptionNet and original XceptionNet architecture.   

%training loss
Figure 2(b) shows the training loss for the models over the epochs. Training loss for optimized XceptionNet and optimized XceptionNet with data are higher than the original XceptionNet architecture. Further, both optimized XceptionNet and optimized XceptionNet with data have almost identical training loss. 

In summary, the proposed optimized XceptionNet and optimized XceptionNet with data are smaller in size and take less training time. However, the training loss of optimized XceptionNet and optimized XceptionNet with data has higher loss values compared to the original XceptionNet architecture. We plan to perform more experimentation in this regard. 

\begin{figure}[!htb]
\centering
\includegraphics[width=0.48\textwidth]{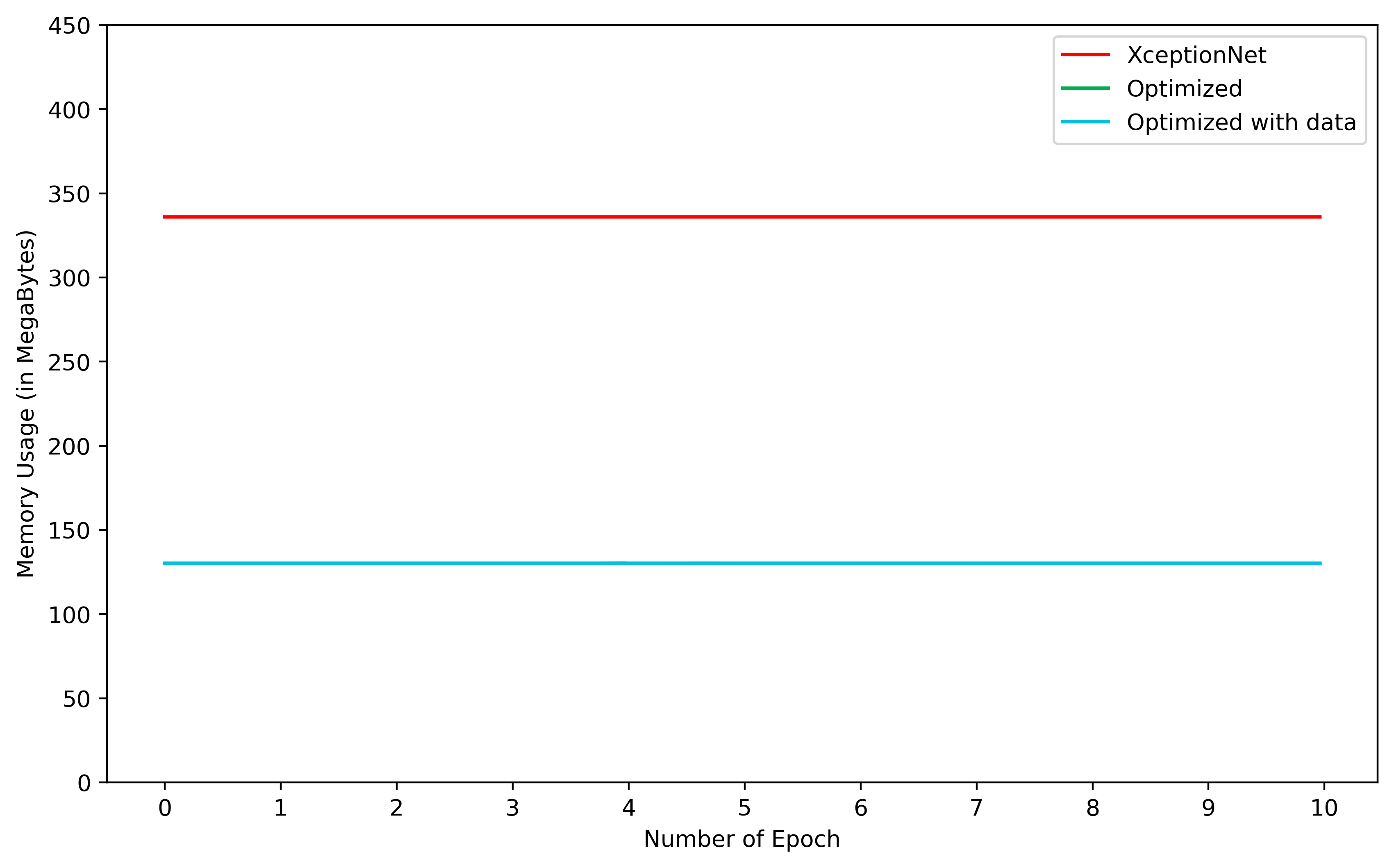}
\caption{Comparison of memory usage among XceptionNet, optimized XceptionNet, and optimized XceptionNet with data}
\label{fig:memory-consumption}
\end{figure}

\subsection{Memory Consumption}
Figure \ref{fig:memory-consumption} shows the memory consumption of original XceptionNet, optimized XceptionNet, and optimized XceptionNet with data. We can observe that the original XceptionNet is taking around 340 megabytes of memory while optimized XceptionNet, and optimized XceptionNet with data are taking around 140 megabytes in main memory. The low memory consumption of the optimized models is no surprise as they are smaller in size.

\subsection{Validation Result}
\begin{figure}[!htb]
\centering
\includegraphics[width=0.48\textwidth]{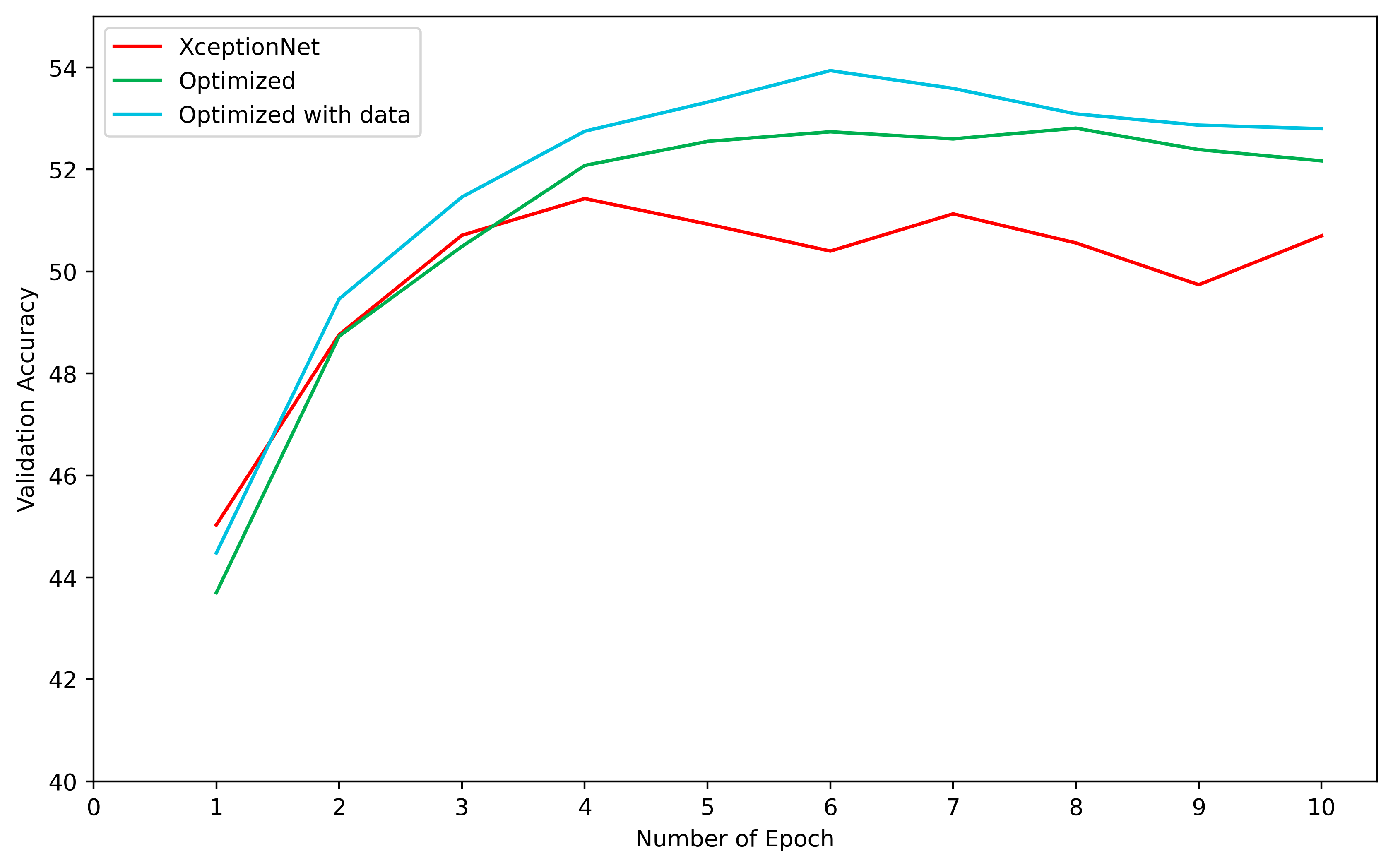}
\caption{Comparison of validation accuracy among XceptionNet, optimized XceptionNet, and optimized XceptionNet with data}
\label{fig:validation-accuracy}
\end{figure}

Figure \ref{fig:validation-accuracy} shows the validation accuracy of the XceptionNet, optimized XceptionNet, and optimized XceptionNet with data. The validation performance of XceptionNet is better than the other two models at epoch one. Optimized XceptionNet matches the accuracy of XceptionNet at epoch two, while optimized XceptionNet with data produces better accuracy than them. Similar performance can also be observed in epoch three. However, from epoch 4 onwards optimized XceptionNet and optimized XceptionNet with data produced better accuracy than XceptionNet, while Optimized XceptionNet with data always produced a better accuracy score than optimized XceptionNet. Optimized XceptionNet with data was growing exponentially from epoch one till epoch six, however from epoch seven onwards the accuracy stopped growing exponentially and converged into a more linear line. Although the accuracy of optimized XceptionNet with data is always better than XceptionNet over the ten each. 

\begin{figure}[!htb]
\centering
\includegraphics[width=0.48\textwidth]{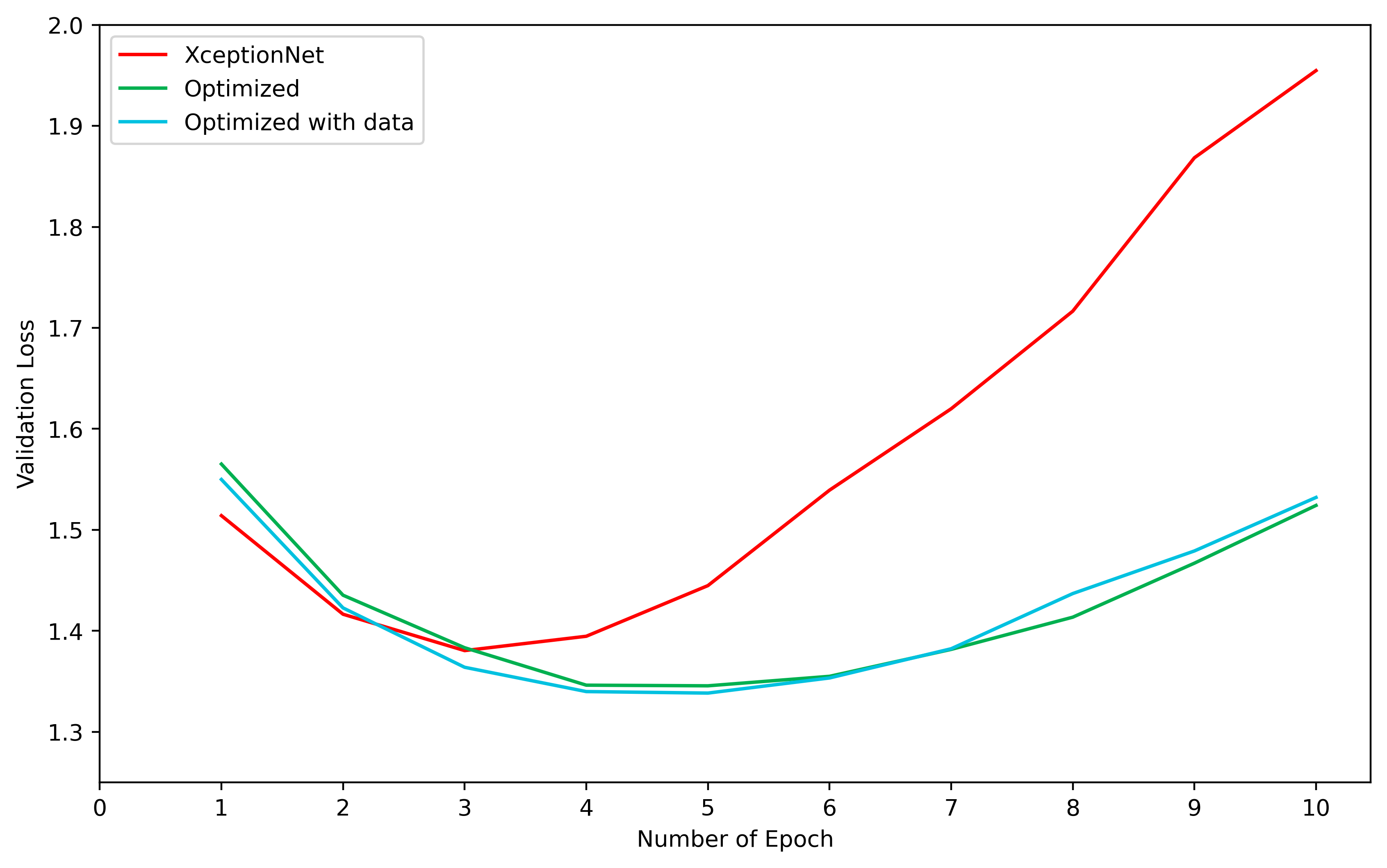}
\caption{Comparison of validation loss among XceptionNet, optimized XceptionNet, and optimized XceptionNet with data}
\label{fig:validation-loss}
\end{figure}

Figure \ref{fig:validation-loss} shows the validation loss of the XceptionNet, optimized XceptionNet, and optimized XceptionNet with data. XceptionNet has the lowest validation loss compared to optimized XceptionNet and optimized XceptionNet with data. The validation loss of all the models decreases significantly in epoch two and epoch three. However, from epoch four onwards validation loss of XceptionNet started growing exponentially and continued throughout the ten epoch. On the other hand validation loss of optimized XceptionNet, and optimized XceptionNet with data kept decreasing until epoch four and remained linear until epoch five. From epoch five onwards they started increasing slowly and continued till epoch ten. The gap in validation loss between XceptionNet and the other two proposed models at epoch ten is 0.4, while validation loss optimized XceptionNet, and optimized XceptionNet with data remained almost identical throughout the ten epoch.

\begin{figure}[!htb]
\centering
\includegraphics[width=0.48\textwidth]{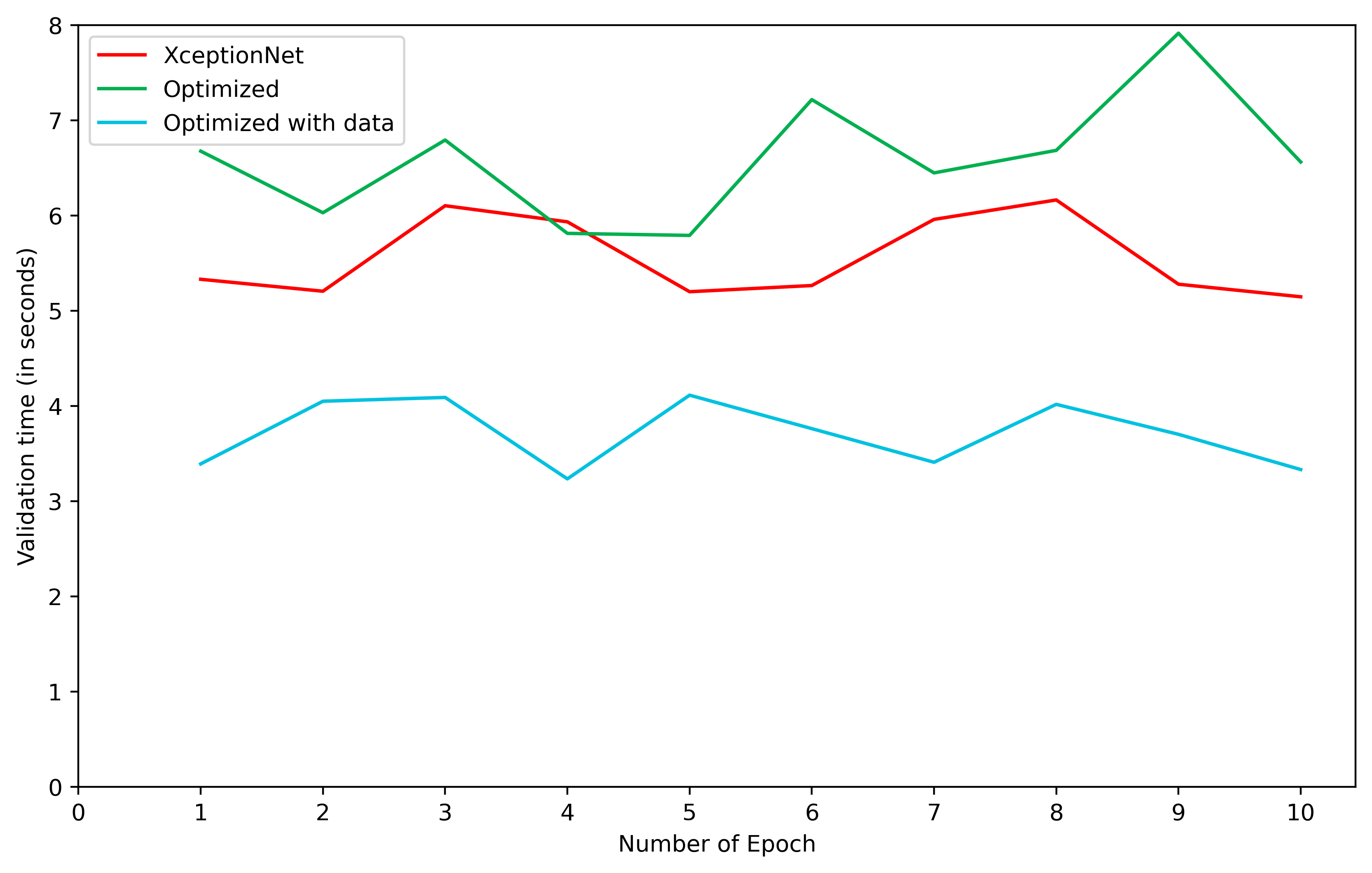}
\caption{Comparison of validation time among XceptionNet, optimized XceptionNet, and optimized XceptionNet with data}
\label{fig:validation-time}
\end{figure}

Figure \ref{fig:validation-time} shows the validation required to validate XceptionNet, optimized XceptionNet, and optimized XceptionNet with data. The figures demonstrated that optimized XceptionNet takes the most time for validation. XceptionNet takes less time for validation compared to optimized XceptionNet, however requires more time than optimized XceptionNet with data. Optimized XceptionNet with data takes around $3-4$ seconds for validation over the ten epochs, while optimized XceptionNet and XceptionNet require $5.5-8$ seconds and  $5-6$ seconds respectively. Optimized XceptionNet takes $0.5-3$ more seconds for validation compared to the original XceptionNet architecture. On the other hand, optimized XceptionNet with data takes $2$ fewer seconds for validation compared to the original XceptionNet architecture.

In summary, the proposed optimized XceptionNet, and optimized XceptionNet with data perform better than the original XceptionNet in terms of training accuracy and loss. Furthermore, optimized XceptionNet takes more time for validation compared to the XceptionNet. However, optimized architecture with data takes less time for validation compared to XceptionNet.

\section{Conclusion}
\label{sec:conclusion}
In this study, we have proposed a modified Xception architecture optimized for deployment on edge devices, addressing the challenges of computational efficiency and resource constraints. We replaced inception modules with depthwise separable convolution layers and incorporated deep residual connections. Our experimental evaluations show improvement in performance and efficiency while reducing parameters, memory usage, and computational load. In the future, we plan to evaluate the proposed model on a set of large-scale object detection datasets to present a fair comparison between the original Xception and proposed optimized Xception architectures. We also plan to experiment with the memory consumption of models in different devices to present a valid comparison.

\bibliographystyle{plain}
\bibliography{bibliography}

\begin{thebibliography}{10}

\bibitem{alex2009learning}
Krizhevsky Alex.
\newblock Learning multiple layers of features from tiny images.
\newblock {\em https://www. cs. toronto. edu/kriz/learning-features-2009-TR. pdf}, 2009.

\bibitem{alzubi2022optimized}
Omar~A Alzubi, Jafar~A Alzubi, Moutaz Alazab, Adnan Alrabea, Albara Awajan, and Issa Qiqieh.
\newblock Optimized machine learning-based intrusion detection system for fog and edge computing environment.
\newblock {\em Electronics}, 11(19):3007, 2022.

\bibitem{Xceptionnet}
Fran{\c{c}}ois Chollet.
\newblock Xception: Deep learning with depthwise separable convolutions.
\newblock {\em CoRR}, abs/1610.02357, 2016.

\bibitem{ding2019tdd}
Runwei Ding, Linhui Dai, Guangpeng Li, and Hong Liu.
\newblock Tdd-net: a tiny defect detection network for printed circuit boards.
\newblock {\em CAAI Transactions on Intelligence Technology}, 4(2):110--116, 2019.

\bibitem{kristiani2020isec}
Endah Kristiani, Chao-Tung Yang, and Chin-Yin Huang.
\newblock isec: An optimized deep learning model for image classification on edge computing.
\newblock {\em IEEE Access}, 8:27267--27276, 2020.

\bibitem{kumar2020energy}
Mohit Kumar, Xingzhou Zhang, Liangkai Liu, Yifan Wang, and Weisong Shi.
\newblock Energy-efficient machine learning on the edges.
\newblock In {\em 2020 IEEE international parallel and distributed processing symposium Workshops (IPDPSW)}, pages 912--921. IEEE, 2020.

\bibitem{9060868}
Wei Yang~Bryan Lim, Nguyen~Cong Luong, Dinh~Thai Hoang, Yutao Jiao, Ying-Chang Liang, Qiang Yang, Dusit Niyato, and Chunyan Miao.
\newblock Federated learning in mobile edge networks: A comprehensive survey.
\newblock {\em IEEE Communications Surveys \& Tutorials}, 22(3):2031--2063, 2020.

\bibitem{merenda2020edge}
Massimo Merenda, Carlo Porcaro, and Demetrio Iero.
\newblock Edge machine learning for ai-enabled iot devices: A review.
\newblock {\em Sensors}, 20(9):2533, 2020.

\bibitem{murshed2021machine}
MG~Sarwar Murshed, Christopher Murphy, Daqing Hou, Nazar Khan, Ganesh Ananthanarayanan, and Faraz Hussain.
\newblock Machine learning at the network edge: A survey.
\newblock {\em ACM Computing Surveys (CSUR)}, 54(8):1--37, 2021.

\bibitem{nikouei2018real}
Seyed~Yahya Nikouei, Yu~Chen, Sejun Song, Ronghua Xu, Baek-Young Choi, and Timothy~R Faughnan.
\newblock Real-time human detection as an edge service enabled by a lightweight cnn.
\newblock In {\em 2018 IEEE International Conference on Edge Computing (EDGE)}, pages 125--129. IEEE, 2018.

\bibitem{nuh2023developing}
Atah Nuh~Mih, Hung Cao, Asfia Kawnine, and Monica Wachowicz.
\newblock Developing a resource-constraint edgeai model for surface defect detection.
\newblock {\em arXiv e-prints}, pages arXiv--2401, 2023.

\bibitem{sudharsan2020edge2train}
Bharath Sudharsan, John~G Breslin, and Muhammad~Intizar Ali.
\newblock Edge2train: A framework to train machine learning models (svms) on resource-constrained iot edge devices.
\newblock In {\em Proceedings of the 10th International Conference on the Internet of Things}, pages 1--8, 2020.

\bibitem{verhelst2020machine}
Marian Verhelst and Boris Murmann.
\newblock Machine learning at the edge.
\newblock {\em NANO-CHIPS 2030: On-Chip AI for an Efficient Data-Driven World}, pages 293--322, 2020.

\bibitem{yazici2018edge}
Mahmut~Taha Yazici, Shadi Basurra, and Mohamed~Medhat Gaber.
\newblock Edge machine learning: Enabling smart internet of things applications.
\newblock {\em Big data and cognitive computing}, 2(3):26, 2018.

\bibitem{zhang2018improved}
Can Zhang, Wei Shi, Xiaofei Li, Haijian Zhang, and Hong Liu.
\newblock Improved bare pcb defect detection approach based on deep feature learning.
\newblock {\em The Journal of Engineering}, 2018(16):1415--1420, 2018.

\end{thebibliography}

\end{document}